\documentclass{article}

   \PassOptionsToPackage{numbers, compress}{natbib}
  \usepackage[preprint]{neurips_2023}

\usepackage[utf8]{inputenc} % allow utf-8 input
\usepackage[T1]{fontenc}    % use 8-bit T1 fonts
\usepackage{hyperref}       % hyperlinks
\usepackage{url}            % simple URL typesetting
\usepackage{booktabs}       % professional-quality tables
\usepackage{amsfonts}       % blackboard math symbols
\usepackage{nicefrac}       % compact symbols for 1/2, etc.
\usepackage{microtype}      % microtypography
\usepackage{xcolor}         % colors
\usepackage{graphicx} 
\usepackage{makecell}
\usepackage{float}
\title{Are Large Language Models Fit For Guided Reading?}

\author{%
  Peter Ochieng \\
  Department of Computer Science\\
  University of Cambridge\\
   \texttt{po304@cam.ac.uk} \\
}

\begin{document}

\maketitle

\begin{abstract}
 This paper looks at the ability of large language models to participate in educational guided reading. We specifically, evaluate their ability to generate meaningful questions from the input text, generate diverse questions both in terms of content coverage and difficulty of the questions and evaluate their ability to recommend part of the text that a student should re-read based on the student's responses to the questions. Based on our evaluation of ChatGPT and Bard, we report that, 
 1) Large language models are able to generate high quality meaningful questions that have high correlation with the input text, 2) They generate diverse question that cover most topics in the input text even though this ability is significantly degraded as the input text increases, 3)The large language models are able to generate both low and high cognitive questions even though they are significantly biased toward low cognitive question, 4) They are able to effectively summarize responses and extract a portion of text that should be  re-read.
\end{abstract}

\section{Introduction}
The recent advancement of natural language processing is currently being exemplified by the  large language model (LLMs) such as GPT-3 \cite{gpt3}, PaLM \cite{Chowdhery2022}, Galactica \cite{Taylor2022} and LLaMA \cite{Touvron2023}. The models have been trained on large amount of text-data and are able to answer questions, generate coherent text and complete most language related tasks. LLMs have been touted to have impact in domains such as climate science \cite{Biswas2023}, health \cite{Hasnain2023} and education \cite{Lund2023}. In education, it has been suggested that they can  be exploited to boost learning in different categories such as in elementary school children, middle and high school children, university students etc \cite{Kasneci2023}. This is line with a long-time goal of AI to develop conversational agents that can support teachers in guiding children through reading material such as reading storybooks \cite{Yao2021} \cite{Zhao2022}. Normally, in reading a text such as children's storybook, a teacher is expected to guide the children through the text and periodically gauge their understanding by posing questions from the text. The key concept in guided reading is the ability to use questions to gauge understanding and encourage deeper thinking about the material being read.  In a teacher led guided reading, apart from gauging understanding, questions can be used to identify children's support needs and enable the teacher to direct attention to the critical content. To achieve the full benefits of guided reading, the teacher is supposed  to ask wide variety of questions ranging from low to high cognitive challenge questions \cite{Blything2020}. Low cognitive challenge questions are constrained to short answers while high cognitive challenge questions require explanations, evaluation or extension of text \cite{Blything2020}.  The use of questions to foster understanding and learning from text is well established across a range of age groups and learning contexts \cite{Blything2020}. It is therefore of interest to gauge the effectiveness of LLMs to perform the tasks involved in guided  reading. For LLMs to be viewed as potential support agents for teachers or even stand-alone tools that can help in guided reading,  they must be able to: generate meaningful questions and answers from the text, generate diverse questions both in terms of content coverage and difficulty of questions  and identify the support needs for the students.  In this work, we investigate the suitability of ChatGPT \footnote{https://openai.com/blog/chatgpt} and Bard \footnote{https://bard.google.com/} to act as a storybook reading guide for children.  Specifically, we evaluate them on the following issues:
\begin{enumerate}
    \item Ability to generate content related  questions and answers from a given input text i.e., its performance in question-answer generation(QAG) task.
    \item Ability to generate both low and high cognitive demand questions.
    \item Ability to generate diverse questions i.e., questions that cover almost all topics in each story.
    \item  Ability to recommend areas that a student needs to focus on based on wrong responses  from the student
    \item Compare their performance to the currently existing AI-based tools for educational question generation.
\end{enumerate}
\section{Related work}
Question generation tools seek to accept input text and generate meaningful questions that are extracted from the text. Existing question generation tools can be categorised into two i.e., rule-based tools and neural based tools \cite{Pan2019}. Rule based  tools such as \cite{Heilman2011} and \cite{sema1} exploit manually crafted rules to extract questions from text. Neural based techniques implement  an end-to-end architecture that follow attention-based  sequence-to-sequence framework \cite{Bahdanau2014}. The sequence-to-sequence frameworks are mainly composed  of two key parts i.e., the encoder which learns a joint representation of the input text and the decoder which generates the questions\cite{Pan2019}. Currently, both the joint representation learning, and  question generation are implemented by attention-based framework \cite{trans2020}. Work in \cite{Du2017} introduced the attention-based sequence-to-sequence architecture to generate question from the input sentence. The encoder was implemented via RNN to accept an input text and learn its representation. Its output is fed into the decoder which generates a related question. The decoder exploits attention mechanism to assign more weight to the most relevant part of the text. Other question asking tools that implement attention-based sequence-to sequence framework include \cite{Duan2017}, \cite{Zhou2017} and \cite{Harrison2018}. Although these neural based tools have a common high-level encoder-decoder structure, their low-level implementations of different aspects of question generation differ significantly \cite{Pan2019}. While these tools are generally developed for question generation, there are tool such as \cite{Yao2021} and \cite{Zhao2022} which target question generation for educational purposes. The use of questions in guided reading has been widely studied \cite{Degener2017} \cite{Ford2015}\cite{Fountas1996}. Questions are used during guided reading to evaluate understanding, encourage deeper thinking about the text and to scaffold understanding of challenging text \cite{Blything2020}. It has been suggested in \cite{Kasneci2023} that large language models can  play a significant role in boosting education of different levels of children. It is therefore important to evaluate them if indeed they are fit for purpose in which they are being deployed in. In this work we evaluate their ability to participate in guided reading and specifically, evaluate their question generation ability. Are they able to follow the general trend of a human teacher in asking question during comprehension reading ?
\section{Question and answer generation (QAG). }
\subsection{Ability to generate meaningful questions.}
LLMs must  demonstrate that they have the potential for an in-depth understanding of a given input text for them to be deployed as reading guides. One indicator of input text's comprehension is the ability to generate meaningful questions and answers from the input text. Formally, given a passage $P$, the model should be able to generate a set of questions $\{q_1,\cdots,q_n\}$ and their respective answers $\{q_1,\cdots,q_n\}$. Both the set of questions and answers should be retrieved from the passage $P$. Perhaps one of the greatest powers of LLMs such as ChatGPT is the ability to respond to questions posed to it on-the-fly. However, it is unclear to what extent they can connect different aspect of input stories to generate both low cognitive questions and questions that require inference to be answered (i.e., high cognitive demand questions). Moreover,how will it exploit the vast amount of knowledge it acquired during training to boost its question asking ability? Further, will it be able to generate answers from the input text without being "confused" by its internal knowledge. We are interested in evaluating the ability of LLMs to ask meaningful questions that can be solely answered from the input story. We are also interested to evaluate  how accurate LLMs are  in answering the questions when solely relying on its understanding of the input text. To do this, we prompt a given LLM to generate questions based on an input story (see fig 1). The generated questions are automatically evaluated  by comparing their semantic similarity to a set of baseline questions. 

 \begin{figure}[ht]
	\centering
\includegraphics[scale=0.4,angle=0]{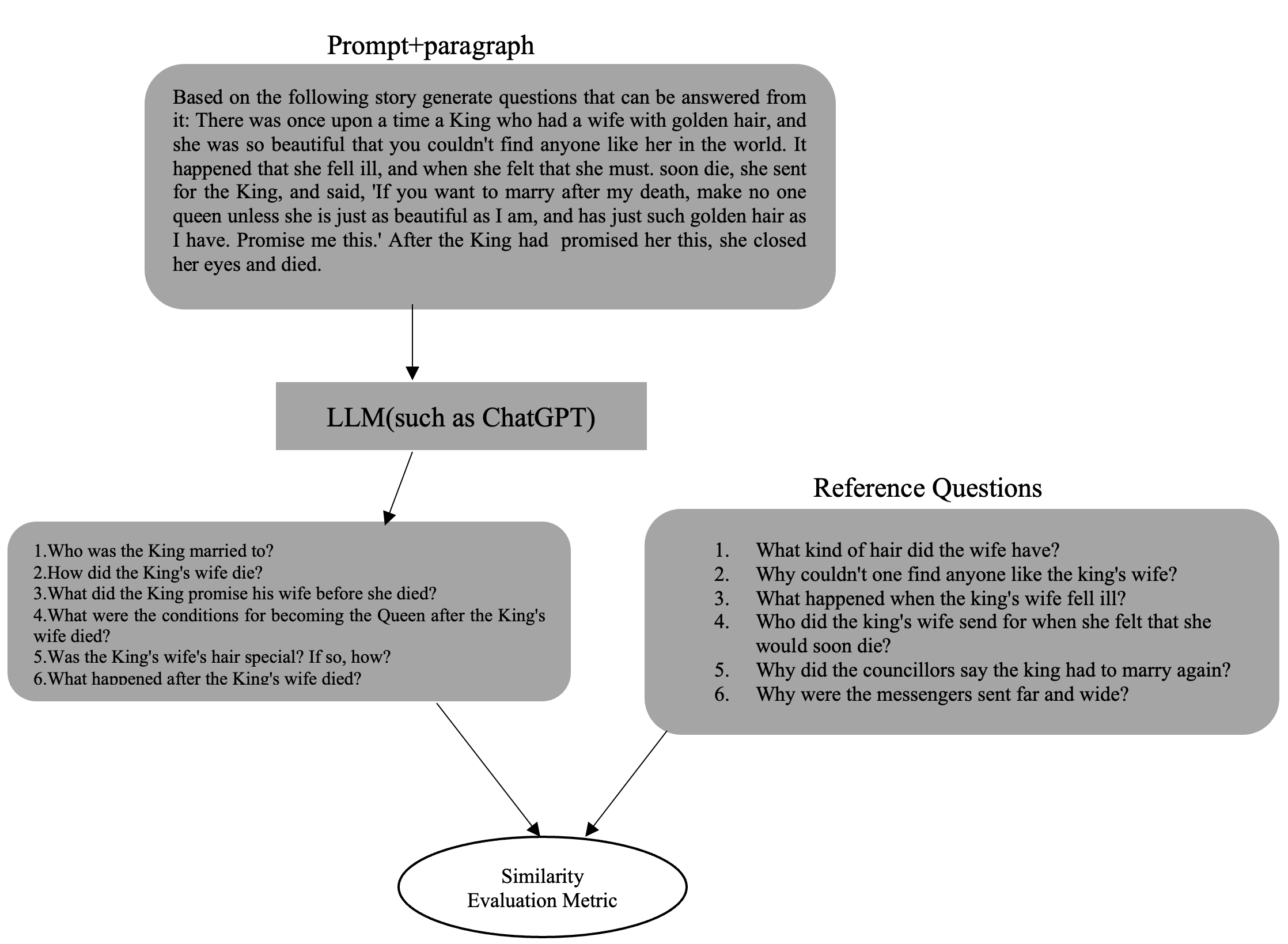}
		\caption{Prompting LLM to generate questions from a given text input.}
	\end{figure}

To evaluate the performance of a given LLM on question generation, we exploit the popularly used metrics in question generation task which include   ROUGE-L \cite{rouge1} and  BERTScore \cite{bertscore1}  and compare the semantic similarity of questions generated with the baseline questions. The similarity between LLM generated  questions and reference questions is evaluated by  concatenating  the generated questions into one sentence and compare it with similarly concatenated reference questions ( see \cite{Zhao2022} which uses a similar approach).

\subsection{Question diversity}
To test the student's understanding of a given text being read, questions must be generated that cover nearly all the sections of the story. We are interested in evaluating the ability of ChatGPT and Bard to generate questions that are not biased towards a given section of the text. Concretely, we are seeking to quantify the variation of the questions being generated by LLMs. We hypothesize that the more diverse the questions are, the more exhaustive the questions cover the different topics in the input text. This will give us an idea on suitability of LLMs to generate questions that cover the whole content being read. In machine learning, several entropy reliant techniques have been proposed to evaluate diversity of dataset. Research in \cite{Salimans2016} proposes Inception Score (IS) to evaluate the synthetic samples generated by a Generative Adversarial model(GAN),$(G)$\cite{Goodfellow2020}. The intuition behind IS is that the conditional label distribution $P(y\mid x)$ of images generated by  GAN is projected to have low label entropy i.e., the generated images belong to few classes while  entropy across the images should be high that is the marginal distribution $\int p(y\mid x=G(z)) dz$ should have high entropy. To capture this intuition, they suggest the metric in equation 1. 
\begin{equation}
exp(\mathbb{E}_x{KL}(p(z\mid x)\parallel p(y)))
\end{equation}
Another popular metric that provides a measure of diversity on synthetically generated samples is the Frechet Inception Distance (FID) score \cite{Heusel2017}. This is a metric that considers location and ordering of the data along the function space.  Taking the activations of the penultimate layer of a given model to represent features of a given dataset $x$ and considering only the first two moments i.e., the mean and covariance, the FID assumes that the coding units of the model $f(x)$ follow a multi-dimensional Gaussian, therefore have maximum entropy distribution for a given mean and covariance. If the model $f(x)$,generates embeddings with mean and covariance (m,C) given the synthetic data $p(.)$  and $(m_w,C_w)$ for real data $p_w(.)$, then FID is defined as:
\begin{equation}
    d^2((m,C),(m_w,C_w)=||m-m_w||_2^2+Tr(C+c_w-2(CC_w)^{1/2})
\end{equation}
Other metrics that have been proposed to evaluate diversity, include precision and recall metrics \cite{Sajjadi2018} \cite{Simon2019}. One major problem with these metrics is that they assume existence of  reference samples where the generated samples can be compared to \cite{Simon2019}. In our case, we seek to evaluate the diversity of the questions generated without comparing to any reference questions. To achieve this, we use Vendi score (VS), a metric proposed for diversity evaluation in the absence of reference data. VS is defined as 
\begin{equation}
    VS(X)=exp(-\sum_i^s \lambda_i log \lambda_i)
\end{equation}
Here $X=\{x_i,\cdots,x_n\}$ is the input data whose diversity is to be evaluated. $\lambda_i, i=\{1,\cdots,n\}$ are the eigenvalues of a positive semidefinite matrix $K/n$ whose entries are $K_{ij}=k(x_i,x_j)$ where $k$ is a positive semidefinite similarity function with $k(x,x)=1$ for all $x$. This  metric which is like effective rank \cite{Roy2007} seeks to quantify the geometric transformation induced by performing a linear mapping  of a vector $x$ from a vector space $R^n$ to $R^m$ by a matrix $A$ i.e $Ax$. Normally, the number of dimensions retained by a linear  transformation $Ax$ is  captured by the rank of the matrix $A$. The rank however is silent on the shape induced by the transformation. Effective rank introduced by \cite{Roy2007} seeks to capture the shape that results due to the linear mapping. Effective rank therefore can be used to capture the spread of data hence ideal to measure diversity. To compute the diversity of questions generated by the two LLMs, we execute the following steps:
\begin{enumerate}
    \item Prompt the LLM to generate a set of questions $Q$ given an input text.
    \item Replicate the set $Q_1=\{q_1,\cdots,q_n\}$  to get a copy of the questions $Q_2=\{q_1,\cdots,q_n\}$. Designate $Q_1$ as the reference set and $Q_2$ as the candidate set.
    \item Pass the set  $Q_1$ and  $Q_2$ through the BERTScore \footnote{\url{https://github.com/Tiiiger/bert_score}}to extract the cosine similarity matrix  $K$.
    \item Use the VS package\footnote{\url{https://github.com/vertaix/Vendi-Score}} to extract the VS score diversity value.
    \item Compare the VS score diversity value to human generated diversity score
\end{enumerate}
\subsubsection{Human generated diversity score}
We engage independent human evaluators to annotate each reference question by attaching a sub-topic they think a given question is addressing in each storybook. We then aggregate all the subtopics generated by the human-annotators. We adopt an annotation by majority. We calculate diversity score of all the storybooks by computing the average number of all the sub-topics generated from all the storybooks' questions i.e., average diversity=$\frac{N}{B}$ i.e., total number of subtopics generated $N$ divided by total number of books $B$. Intuitively average diversity represents the number of different sub-topics  per storybook.

\subsection{Ability to generate questions that differ in difficulty}
On top of generating questions that cover the whole content, it is desirable for LLMs to generate wide range of questions from low to high cognitive challenge questions. This will make students to answer questions that address  all the cognitive domains. Low cognitive question mostly requires short answers which require affirmation or not (e.g.  "Did the queen have golden hair ?").  Conversely high cognitive challenge questions require explanations, evaluations and some speculations on the extension of text \cite{Blything2020}(e.g., "Why did the king's advisors fail to find the right wife for him ?"). The purpose of low cognitive challenge questions is to evaluate the basic understanding of the text by the students and ease them into the study interaction \cite{Blything2020}. However, they  have the potential of promoting over-dominance of teachers . On the other hand, high cognitive questions foster greater engagement of the students, generate inferential responses and promote better comprehension of the reading material \cite{Blything2020}. In \cite{Blything2020}, the two categories of questions are differentiated by exploiting the syntactic structure of the question. The  high cognitive challenge questions  are signalled  by  wh-word i.e., questions that use words such as  what, why, how, or when. These questions are also composed of a continuum of high to low challenge questions. Specifically, wh-pronoun and wh-determinat questions that start with who, whom, whoever, what, which, whose, whichever and whatever require low challenge literal responses. However, the wh-adverb questions such how, why, where, and when  are more challenging since they require more abstract and inferential responses involving explanation of causation and evaluation (e.g., “Why was the king sad?”; “How did the king's daughter receive the news of her marriage?”). The low cognitive challenge questions are generally non wh-word questions. It has been suggested in \cite{Degener2017}\cite{Zucker2010} that teachers should generally seek to ask high challenge questions as opposed to low challenge questions whenever possible.   In our case we seek to establish the types of questions preferred by the two LLMs based on their level of challenge. To evaluate this, we adopt three categories of questions i.e., confirmative, explicit and implicit. Explicit are non-confirmative questions that require low challenge literal responses i.e., where answers can be retrieved from the text without much inference  while implicit are questions that require inferential responses. To evaluate the type of questions generated, we employ  the following steps:
\begin{enumerate}
    \item  Given a paragraph of text we prompt the LLM to generate questions based on the text.
    \item  We provide the questions generated to human evaluators and ask them to read the text and answer the questions. 
    \item Human evaluators then  annotate each question whether it is confirmative , explicit and implicit.
    \item For each question, we select the most popular annotation i.e., annotation where most evaluators agree.
    \item We compute percentages of each category.
\end{enumerate}
\subsection{Ability to recommend section of text}
Based on the responses  to the teacher's questions, the teacher can detect students' weaknesses and their demands \cite{Habib2016}. While it was difficult to design an evaluation on LLMs that can uncover students' weaknesses based on their responses,  we resorted to evaluate their ability to recommend part of text where the student needs to re-read based on the responses provided to the questions. Basically, we evaluate the ability of a LLM to detect part of the text that the student did not understand. This we  believe plays some part  in diagnosing student's need. To perform our evaluation, we execute the following steps:
\begin{enumerate}
    \item We pass a three-paragraph text to a large language model and prompt it to generate questions from the text.
    \item We annotate the generated questions based on the paragraph in which they were extracted from i.e $<p_i,q_i>$ where $p_i$ is the paragraph $i=1,2,3$ and $q_i$  are the questions $i=1,\cdots,n$, linked to paragraph $p_i$.
    \item We prompt the large language model to answer all the questions generated. We then deliberately alter a set of answers belonging to questions from a  given paragraph to be wrong. All the answers from other two paragraphs remain as generated by the LLM.
    \item We then prompt LLM to evaluate all the answers( for all the questions) generated in the previous step.
    \item We prompt LLMs to suggest the section of the text that the student did not comprehend based on the responses in the previous step.
    \item We compute the BERTScore between the recommended text and the paragraph which had all question altered to be wrong.
\end{enumerate}
\section{Experimental setup}
\subsection{Dataset}
We use the FairytaleQA dataset \cite{Yao2021} to perform evaluation. The dataset contains 278 annotated storybooks which is composed of 232 storybooks for training a given question generation model, 23 books for testing and 23 for validation. Each book contains multiple paragraphs. Each paragraph of a book is annotated with several educational question-answer pairs. The questions have different types of annotation with our main concern being annotations linked to the difficulty attached to answering the questions. The question difficulty is annotated as either implicit (high challenge questions) or explicit (low challenge questions). The dataset covers storybooks for  children between kindergarten and eighth grade.
\subsection{Baseline models}
For question generation, we compare the performance of ChatGPT and Bard with selected models that were specifically developed for questions generations and were evaluated in FairytaleQA dataset. The models include:\\
\textbf{QAG} \cite{Yao2021}: The model relies on  semantic role labelling to identify entities and events. The identified entities and events are then used to generate questions. We use the results reported in \cite{Zhao2022}. The questions used in evaluation include top 10/5/3/2/1 generated questions by QAG,  labeled here as (top 10), QAG (top 5), QAG (top 3), QAG (top 1) respectively.\\
\textbf{E2E}: This model is proposed in \cite{Zhao2022} where  one BART-large model is trained on FairytaleQA dataset to generate questions. \\
\textbf{BERT based model}: \cite{Zhao2022}. This model adapts and trains BERT to learn question type distribution. Using the question type distribution as a control signal, they then  train BART summarization model using FairytaleQA dataset question-answer annotation to learn the event-centric summary generation task. \\
\subsection{Human evaluators}
Human evaluators used in the study were recruited through post-graduate social media page in Kenya. An advert was posted on the page where 109 Msc. and PhD students responded positively indicating their willingness to participate. Out of these we selected 36 students who are doing research in various topics in NLP. Each student was paid \$30 once the annotation task was completed. We conducted one day  online training via zoom on the annotation process.

\section{Quality of questions generated}
The results of the similarity of the questions generated by Bard and ChatGPT as compared to FairytaleQA dataset  human annotated questions are shown in table 2. We only used the 46 storybooks which are contained in the test and validation set. This is to enable direct comparison with baseline models which also use the two sets. Table 2 also shows the performance of the baseline models. Based on the results, both ChatGPT and Bard register a slight performance advantage in their f-measure values. The lack of significant advantage of LLMs  over the baseline models is surprising given the high quality of questions they generated  based on human evaluation. We hypothesise that the evaluation based on matching of similar tokens (used in ROUGE-L )  may not be ideal for this set-up since both ChatGPT and Bard were trained using very large text datasets that have  extensive vocabularies, therefore they have a wide space of vocabularies to use while asking a question. Hence some of their vocabulary’s choices  may not be present in the reference questions. Table 1 shows some examples where the questions are  semantically similar, but Rouge-L values are low. Further unlike baseline models,  ChatGPT and Bard were not trained exclusively on  FairytaleQA dataset question-answer pair, hence their style of asking questions may be significantly different from the reference questions.
    \begin{table}[ht]
\centering
\caption{Sample questions where ChatGPT uses diverse vocabulary.} 
\centering
\resizebox{\textwidth}{!}{\begin{tabular}{|l|l|l|l|l|}
    \hline
    
    \textbf{Reference Questions}&\textbf{ChatGPT generated question}& \textbf{Precision}& \textbf{Recall}& \textbf{f-measure}\\ \hline
     \multicolumn{1}{|l|}{What kind of hair did the wife have ?}& {What was the Queen's hair color?}&0.2857&0.2500&0.2667\\ \hline
     \multicolumn{1}{|l|}{Why did the councillors say the king had to marry again?}&Why was the King advised to re-marry ?&0.4545&0.62500&0.5263\\ \hline
       \multicolumn{1}{|l|}{Who did the king's wife send for when she felt that she would soon die}& Who did the Queen send for when she fell sick ?&0.4375&0.700&0.5384\\ \hline
       
\end{tabular}}
\end{table}
However, the superiority of LLMs is demonstrated when using BERTScore metric. Here, they outperform the baseline models in precision, recall and f-measure values. Their superior performance demonstrates that LLMs are able to generate meaningful educational questions that mimic how a human-teacher would ask questions.
\begin{table}[ht]
\centering
\caption{Evaluation of quality of questions generated by ChatGPT and Bard.}

\begin{tabular}{|r|r|r|r|}
    \hline
    \multicolumn{4}{|c|}{\textbf{Rouge-L results on validation/test dataset} } \\ \hline  
    \textbf{ Model} & \textbf{Precision} &\textbf{ Recall} & \textbf{f-measure}\\ \hline
     \multicolumn{1}{|l|}{E2E}& 16.32/15.76& 36.21/35.89 &20.29/19.73  \\ \hline
    \multicolumn{1}{|l|}{QAG(top-1)}& 34.58/32.33& 19.56/19.69 & 22.88/22.29  \\ \hline
    \multicolumn{1}{|l|}{QAG(top-2)}& 28.45/26.58 & 30.51/30.34&26.76/25.67 \\ \hline
    \multicolumn{1}{|l|}{QAG(top-3)} & 24.29/22.74 & 36.80/36.31 & 26.67/25.50  \\ \hline
     \multicolumn{1}{|l|}{QAG(top-5)} & 20.38/19.25&43.45/43.04 & 25.55/24.53 \\ \hline
     \multicolumn{1}{|l|}{QAG(top-10)} & 18.12/17.26 & 46.57/47.04 &24.05/23.34  \\ \hline
     \multicolumn{1}{|l|}{BERT based model\cite{Zhao2022} }& 33.49/37.50 & 37.50/31.54& 31.81/30.58  \\ \hline
     \multicolumn{1}{|l|}{ChatGPT} & 31.21/33.45& 39.78/42.33 & 34.98/37.36\\ \hline
     \multicolumn{1}{|l|}{Bard} & 21.62/26.12&38.09/45.89 & 27.58/33.29 \\ \hline
    \multicolumn{4}{|c|}{\textbf{BERTScore results on validation/test dataset}} \\ \hline
   \multicolumn{1}{|l|}{E2E}& 88.55/88.39& 84.25/84.07 &86.32/86.15  \\ \hline
    \multicolumn{1}{|l|}{QAG(top-1)}& 85.99/86.23& 87.76/87.70 & 86.84/86.94  \\ \hline
    \multicolumn{1}{|l|}{QAG(top-2)}&88.30/88.105/ & 87.45/87.02&87.86/87.54 \\ \hline
    \multicolumn{1}{|l|}{QAG(top-3)} &88.66/88.46/& 86.63/86.29 & 87.61/87.34\\ \hline
     \multicolumn{1}{|l|}{QAG(top-5)} & 88.83/88.62&85.71/85.40& 87.22/86.96 \\ \hline
     \multicolumn{1}{|l|}{QAG(top-10)} & 88.73/88.48& 85.03/84.72 &86.81/86.54  \\ \hline
     \multicolumn{1}{|l|}{BERT based model\cite{Zhao2022}} & 89.15/88.62 & 88.86/89.30& 88.98/88.93  \\ \hline
     \multicolumn{1}{|l|}{ChatGPT} & 96.92/96.31&95.03/96.01& 95.96/96.15\\ \hline
     \multicolumn{1}{|l|}{Bard} &  97.12/93.31&95.43/96.34& 96.27/94.80\\ \hline
\end{tabular}
\end{table}
\section{Question diversity on topic coverage}
The FairytaleQA dataset  has 23 books to be used for model  testing and 23  for validation. We use these 46 books to evaluate if ChatGPT and Bard can cover all the subtopics of a storybook while asking questions. To do this we use the evaluation criteria described in section 3.2. Question-topic diversity is evaluated on per book basis. For a given book, we generate questions by passing one or merged paragraphs  and a prompt into a large language model. We iterate through all the paragraphs of the book and aggregate generated questions into a set Q. We then replicate the questions to get another set $Q^{'}$. BERTScore is then exploited to generate the similarity matrix $K$. The established matrix $K$ is  used to compute VS value for that book. To get an  idea of how a given large language model generates diverse questions, we average the VS values over the 46 books. We also engage human annotators to annotate FairytaleQA reference questions based on the 46 books. The annotators are supposed to label each question with the sub-topic that the question addresses. We do not restrict the possible topics but allow annotators to come up with their own based on reading the storybooks excerpts and the questions. Annotations are adopted by majority. From human evaluators, we generated an average diversity score of 66.8 from the 46 excerpts of the storybooks. The question-topic diversity results are shown in table 3,4 and 5. We increase the input text by varying the number of paragraphs from 1 to 3.
From the results in table1, when one paragraph is used as input, taking human annotation as the baseline, ChatGPT generates questions that cover slightly above 61 sub-topics in each storybook. This represents 91.9\% of the total sub-topics. Bard large language model covers 64 sub-topics per given storybook. This represents 96.1\% of the subtopics. In general, both LLMs generate questions that exhibit high diversity when compared to human generated diversity. However, as the size of input increases from 1 to 2  merged paragraphs, the diversity score of both LLMs drop to 54.4\% and 56.2\% for ChatGPT and Bard respectively. A further increase of input to 3 merged paragraphs reduces the diversity score to 49.4\% and 48.2\% for ChatGPT and Bard respectively. This is an indication LLMs are still limited on the amount of content that they can effectively handle.
\begin{table}[ht]
\centering
\caption{sub-topic diversity evaluation of LLMs with a one paragraph input.}
\begin{tabular}{|r|r|}
    \hline
    \multicolumn{2}{|c|}{\textbf{Rouge-L results on validation/test dataset} } \\ \hline  
    \textbf{Model}&\textbf{VS} \\ \hline
     \multicolumn{1}{|l|}{Human Evaluators}& 66.8\\ \hline

       \multicolumn{1}{|l|}{ChatGPT} & 61.4\\ \hline
     \multicolumn{1}{|l|}{Bard} & 64.2\\ \hline
    
\end{tabular}
\end{table}
\begin{table}[ht]
\centering
\caption{sub-topic diversity evaluation of LLMs with  two merged paragraphs as input.}
\begin{tabular}{|r|r|}
    \hline
    \multicolumn{2}{|c|}{\textbf{Rouge-L results on validation/test dataset} } \\ \hline  
    \textbf{Model}&\textbf{VS} \\ \hline
     \multicolumn{1}{|l|}{Human Evaluators}& 66.8\\ \hline

       \multicolumn{1}{|l|}{ChatGPT} & 54.4\\ \hline
     \multicolumn{1}{|l|}{Bard} & 56.2\\ \hline
    
\end{tabular}
\end{table}

\begin{table}[ht]
\centering
\caption{sub-topic diversity evaluation LLMs with  three merged paragraphs as input.}
\begin{tabular}{|r|r|}
    \hline
    \multicolumn{2}{|c|}{\textbf{Rouge-L results on validation/test dataset} } \\ \hline  
    \textbf{Model}&\textbf{VS} \\ \hline
     \multicolumn{1}{|l|}{Human Evaluators}& 66.8\\ \hline

       \multicolumn{1}{|l|}{ChatGPT} & 49.4\\ \hline
     \multicolumn{1}{|l|}{Bard} & 48.2\\ \hline
    
\end{tabular}
\end{table}
\section{Question diversity based on difficulty results}
Based on the results in table 6, baseline questions contain 53.22 \% low challenge questions( i.e., conformative  and explicit questions) while ChatGPT and Bard generate 70.26\% and 73.38\% low challenge questions respectively. This is approximately a 20 \% deviation from the baseline. While research in \cite{Blything2020} confirms that the  current trend of human teachers is to ask more low challenge questions as compared to  high cognitive questions, LLMs significantly over-generate low challenge questions as compared to the baseline. Therefore, there is need to moderate the questions to reflect acceptable human-teachers way of asking questions.
\begin{table}[H]
\centering
\caption{Cognitive level of questions asked by LLMs}
\begin{tabular}{|r|r|r|r|r|}
    \hline
    \multicolumn{5}{|c|}{\textbf{What Questions Do LLMs Ask?} } \\ \hline  
    \textbf{Question Source}& \textbf{total questions}&\textbf{confirmative} &\textbf{explicit} &\textbf{implicit}\\ \hline
     \multicolumn{1}{|l|}{Baseline}& 2030& 0&1121&909\\ \hline
     \multicolumn{1}{|l|}{ChatGPT}& 3006& 109&2003&894\\ \hline
      \multicolumn{1}{|l|}{Bard}& 3546& 97&2505&947\\ \hline

\end{tabular}
\end{table}

\section{Text recommendation}
Here, we report the results of text recommendation of ChatGPT and Bard based on its ability to evaluate the responses and select part of text that the student did not understand. Both language models perform highly in text recommendation. This is an indication that they have the ability to summarize the student's responses and extract section of the story that the student needs to re-read.
\begin{table}[H]
\centering
\caption{LLMs' ability to recommend relevant text.}
\begin{tabular}{|r|r|r|r|}
    \hline
    \multicolumn{4}{|r|}{\textbf{BERTScore values} } \\ \hline  
    \textbf{LLM}& \textbf{Precsion}&\textbf{Recall} &\textbf{f-measure} \\ \hline
     \multicolumn{1}{|l|}{ChatGPT}& 98.22& 99.1&98.65\\ \hline
      \multicolumn{1}{|l|}{Bard}& 97.3& 98.7&97.99\\ \hline

\end{tabular}
\end{table}
\section{Limitations}
In our evaluation, we used a dataset that has storybooks that cover children from Kindergarten to eighth grade. The evaluation needs to be extended to storybooks covering more advanced levels of students. This will evaluate the ability of LLMs to generate high levels questions no-matter the cognitive level of the input text. A good part of the study relied on restricted number of human evaluators. While we  ensured that we eliminated detected biases, the study can be replicated by increasing the number of human evaluators and choosing a more diverse population in-terms of location and level of education. There is  also an opportunity to re-look at the text semantic similarity comparison metrics where one model is able to generate sentences based of a large sample space as compared to restricted vocabulary of the reference text. Learning is a complex process that is influenced by many parameters, therefore the use of non-human teachers on students needs to be thoroughly investigated before deploying LLMs based tools.

\section{Conclusion}
The results presented in this paper demonstrates that large language models have made significant progress in text comprehension and have the potential of being exploited by teachers as tools to assist during guided reading process. However, further validation of the results is needed by evaluating LLMs using diverse datasets, performing an in-depth analysis of the type of questions generated and how they directly correlate with human-teacher questions. Its social impact on teachers and students needs to be evaluated before they are deployed as reading guide assisting tools.

\bibliographystyle{IEEEtran}
\bibliography{mybibfile}  

\begin{thebibliography}{10}
\providecommand{\url}[1]{#1}
\csname url@samestyle\endcsname
\providecommand{\newblock}{\relax}
\providecommand{\bibinfo}[2]{#2}
\providecommand{\BIBentrySTDinterwordspacing}{\spaceskip=0pt\relax}
\providecommand{\BIBentryALTinterwordstretchfactor}{4}
\providecommand{\BIBentryALTinterwordspacing}{\spaceskip=\fontdimen2\font plus
\BIBentryALTinterwordstretchfactor\fontdimen3\font minus
  \fontdimen4\font\relax}
\providecommand{\BIBforeignlanguage}[2]{{%
\expandafter\ifx\csname l@#1\endcsname\relax
\typeout{** WARNING: IEEEtran.bst: No hyphenation pattern has been}%
\typeout{** loaded for the language `#1'. Using the pattern for}%
\typeout{** the default language instead.}%
\else
\language=\csname l@#1\endcsname
\fi
#2}}
\providecommand{\BIBdecl}{\relax}
\BIBdecl

\bibitem{gpt3}
\BIBentryALTinterwordspacing
T.~B. Brown, B.~Mann, N.~Ryder, M.~Subbiah, J.~Kaplan, P.~Dhariwal,
  A.~Neelakantan, P.~Shyam, G.~Sastry, A.~Askell, S.~Agarwal, A.~Herbert-Voss,
  G.~Krueger, T.~Henighan, R.~Child, A.~Ramesh, D.~M. Ziegler, J.~Wu,
  C.~Winter, C.~Hesse, M.~Chen, E.~Sigler, M.~Litwin, S.~Gray, B.~Chess,
  J.~Clark, C.~Berner, S.~Mccandlish, A.~Radford, I.~Sutskever, and D.~Amodei,
  ``Language models are few-shot learners,'' \emph{Advances in neural
  information processing systems}, vol.~33, pp. 1877--1901., 2020. [Online].
  Available: \url{https://commoncrawl.org/the-data/}
\BIBentrySTDinterwordspacing

\bibitem{Chowdhery2022}
\BIBentryALTinterwordspacing
A.~Chowdhery, S.~Narang, J.~Devlin, M.~Bosma, G.~Mishra, A.~Roberts, P.~Barham,
  H.~W. Chung, C.~Sutton, S.~Gehrmann, P.~Schuh, K.~Shi, S.~Tsvyashchenko,
  J.~Maynez, A.~Rao, P.~Barnes, Y.~Tay, N.~Shazeer, V.~Prabhakaran, E.~Reif,
  N.~Du, B.~Hutchinson, R.~Pope, J.~Bradbury, J.~Austin, M.~Isard, G.~Gur-Ari,
  P.~Yin, T.~Duke, A.~Levskaya, S.~Ghemawat, S.~Dev, H.~Michalewski, X.~Garcia,
  V.~Misra, K.~Robinson, L.~Fedus, D.~Zhou, D.~Ippolito, D.~Luan, H.~Lim,
  B.~Zoph, A.~Spiridonov, R.~Sepassi, D.~Dohan, S.~Agrawal, M.~Omernick, A.~M.
  Dai, T.~S. Pillai, M.~Pellat, A.~Lewkowycz, E.~Moreira, R.~Child, O.~Polozov,
  K.~Lee, Z.~Zhou, X.~Wang, B.~Saeta, M.~Diaz, O.~Firat, M.~Catasta, J.~Wei,
  K.~Meier-Hellstern, D.~Eck, J.~Dean, S.~Petrov, and N.~Fiedel, ``Palm:
  Scaling language modeling with pathways,'' \emph{arXiv preprint
  arXiv:2204.02311}, 4 2022. [Online]. Available:
  \url{http://arxiv.org/abs/2204.02311}
\BIBentrySTDinterwordspacing

\bibitem{Taylor2022}
\BIBentryALTinterwordspacing
R.~Taylor, M.~Kardas, G.~Cucurull, T.~Scialom, A.~Hartshorn, E.~Saravia,
  A.~Poulton, V.~Kerkez, and R.~Stojnic, ``Galactica: A large language model
  for science,'' \emph{arXiv preprint arXiv:2211.0908}, 11 2022. [Online].
  Available: \url{http://arxiv.org/abs/2211.09085}
\BIBentrySTDinterwordspacing

\bibitem{Touvron2023}
\BIBentryALTinterwordspacing
H.~Touvron, T.~Lavril, G.~Izacard, X.~Martinet, M.-A. Lachaux, T.~Lacroix,
  B.~Rozière, N.~Goyal, E.~Hambro, F.~Azhar, A.~Rodriguez, A.~Joulin,
  E.~Grave, and G.~Lample, ``Llama: Open and efficient foundation language
  models,'' \emph{arXiv preprint arXiv:2302.13971.}, 2 2023. [Online].
  Available: \url{http://arxiv.org/abs/2302.13971}
\BIBentrySTDinterwordspacing

\bibitem{Biswas2023}
S.~S. Biswas, ``Potential use of chat gpt in global warming,'' \emph{Annals of
  Biomedical Engineering}, 2023.

\bibitem{Hasnain2023}
\BIBentryALTinterwordspacing
M.~Hasnain, ``Chatgpt applications and challenges in controlling monkey pox in
  pakistan,'' \emph{Annals of Biomedical Engineering}, 5 2023. [Online].
  Available: \url{https://link.springer.com/10.1007/s10439-023-03231-z}
\BIBentrySTDinterwordspacing

\bibitem{Lund2023}
\BIBentryALTinterwordspacing
B.~D. Lund and T.~Wang, ``Chatting about chatgpt: How may ai and gpt impact
  academia and libraries?'' \emph{Library Hi Tech News.}, 2023. [Online].
  Available: \url{https://ssrn.com/abstract=4333415}
\BIBentrySTDinterwordspacing

\bibitem{Kasneci2023}
E.~Kasneci, K.~Sessler, S.~K. Uchemann, M.~Bannert, D.~Dementieva, F.~Fischer,
  U.~Gasser, G.~Groh, S.~G. Unnemann, S.~Krusche, G.~Kutyniok, T.~Michaeli,
  C.~Nerdel, J.~U. Pfeffer, O.~Poquet, M.~Sailer, A.~Schmidt, T.~Seidel,
  M.~Stadler, J.~Weller, J.~Kuhn, and G.~Kasneci, ``Chatgpt for good? on
  opportunities and challenges of large language models for education,''
  \emph{Learning and Individual Differences}, vol. 103, 2023.

\bibitem{Yao2021}
\BIBentryALTinterwordspacing
B.~Yao, D.~Wang, T.~Wu, Z.~Zhang, T.~J.-J. Li, M.~Yu, and Y.~Xu, ``It is ai's
  turn to ask humans a question: Question-answer pair generation for children's
  story books,'' \emph{arXiv preprint arXiv:2109.03423}, 9 2021. [Online].
  Available: \url{http://arxiv.org/abs/2109.03423}
\BIBentrySTDinterwordspacing

\bibitem{Zhao2022}
Z.~Zhao, Y.~Hou, D.~Wang, M.~Yu, C.~Liu, and X.~Ma, ``Educational question
  generation of children storybooks via question type distribution learning and
  event-centric summarization,'' \emph{arXiv preprint arXiv:2203.14187},
  vol.~1, pp. 5073--5085, 2022.

\bibitem{Blything2020}
L.~P. Blything, A.~Hardie, and K.~Cain, ``Question asking during reading
  comprehension instruction: A corpus study of how question type influences the
  linguistic complexity of primary school students’ responses,''
  \emph{Reading Research Quarterly}, vol.~55, pp. 443--472, 7 2020.

\bibitem{Pan2019}
L.~Pan, W.~Lei, T.-S. Chua, and M.-Y. Kan, ``Recent advances in neural question
  generation,'' \emph{arXiv preprint arXiv:1905.08949}, 2019.

\bibitem{Heilman2011}
M.~Heilman, ``Automatic factual question generation from text,'' Ph.D.
  dissertation, Carnegie Mellon University, 2011.

\bibitem{sema1}
Y.~Chali and S.~A. Hasan, ``Towards automatic topical question generation,'' in
  \emph{Proceedings of COLING 2012}, 2012, pp. 475--492.

\bibitem{Bahdanau2014}
\BIBentryALTinterwordspacing
D.~Bahdanau, K.~Cho, and Y.~Bengio, ``Neural machine translation by jointly
  learning to align and translate,'' \emph{arXiv preprint arXiv:1409.0473.}, 9
  2014. [Online]. Available: \url{http://arxiv.org/abs/1409.0473}
\BIBentrySTDinterwordspacing

\bibitem{trans2020}
A.~Vaswani, N.~Shazeer, N.~Parmar, J.~Uszkoreit, L.~Jones, A.~N. Gomez, Łukasz
  Kaiser, and I.~Polosukhin, ``Attention is all you need,'' \emph{Advances in
  Neural Information Processing Systems}, vol. 2017-Decem, pp. 5999--6009,
  2017.

\bibitem{Du2017}
X.~Du, J.~Shao, and C.~Cardie, ``Learning to ask: Neural question generation
  for reading comprehension,'' \emph{arXiv preprint arXiv:1705.00106}, 2017.

\bibitem{Duan2017}
N.~Duan, D.~Tang, P.~Chen, and M.~Zhou, ``Question generation for question
  answering,'' in \emph{Proceedings of the 2017 conference on empirical methods
  in natural language processing}, 2017, pp. 866--874.

\bibitem{Zhou2017}
Q.~Zhou, N.~Yang, F.~Wei, C.~Tan, H.~Bao, and M.~Zhou, ``Neural question
  generation from text: A preliminary study,'' in \emph{Natural Language
  Processing and Chinese Computing: 6th CCF International Conference, NLPCC
  2017, Dalian, China, November 8--12, 2017, Proceedings 6}.\hskip 1em plus
  0.5em minus 0.4em\relax Springer, 2018, pp. 662--671.

\bibitem{Harrison2018}
V.~Harrison and M.~Walker, ``Neural generation of diverse questions using
  answer focus, contextual and linguistic features,'' \emph{arXiv preprint
  arXiv:1809.02637}, 2018.

\bibitem{Degener2017}
S.~Degener and J.~Berne, ``Complex questions promote complex thinking,''
  \emph{Reading Teacher}, vol.~70, pp. 595--599, 3 2017.

\bibitem{Ford2015}
M.~P. Ford, \emph{Guided reading: What's new, and what's next}.\hskip 1em plus
  0.5em minus 0.4em\relax Capstone, 2015.

\bibitem{Fountas1996}
I.~C. Fountas and G.~S. Pinnell, \emph{Guided reading: Good first teaching for
  all children.}\hskip 1em plus 0.5em minus 0.4em\relax ERIC, 1996.

\bibitem{rouge1}
C.-Y. Lin, ``Rouge: A package for automatic evaluation of summaries,'' in
  \emph{Text summarization branches out}, 2004, pp. 74--81.

\bibitem{bertscore1}
T.~Zhang, V.~Kishore, F.~Wu, K.~Q. Weinberger, and Y.~Artzi, ``Bertscore:
  Evaluating text generation with bert,'' \emph{arXiv preprint
  arXiv:1904.09675}, 2019.

\bibitem{Salimans2016}
T.~Salimans, I.~Goodfellow, W.~Zaremba, V.~Cheung, A.~Radford, and X.~Chen,
  ``Improved techniques for training gans,'' \emph{Advances in neural
  information processing systems}, vol.~29, 2016.

\bibitem{Goodfellow2020}
I.~Goodfellow, J.~Pouget-Abadie, M.~Mirza, B.~Xu, D.~Warde-Farley, S.~Ozair,
  A.~Courville, and Y.~Bengio, ``Generative adversarial networks,''
  \emph{Communications of the ACM}, vol.~63, pp. 139--144, 10 2020.

\bibitem{Heusel2017}
\BIBentryALTinterwordspacing
M.~Heusel, H.~Ramsauer, T.~Unterthiner, B.~Nessler, and S.~Hochreiter, ``Gans
  trained by a two time-scale update rule converge to a local nash
  equilibrium,'' \emph{Advances in neural information processing systems,},
  vol.~30, 6 2017. [Online]. Available: \url{http://arxiv.org/abs/1706.08500}
\BIBentrySTDinterwordspacing

\bibitem{Sajjadi2018}
M.~S. Sajjadi, O.~Bachem, M.~Lucic, O.~Bousquet, and S.~Gelly, ``Assessing
  generative models via precision and recall,'' \emph{Advances in neural
  information processing systems}, vol.~31, 2018.

\bibitem{Simon2019}
\BIBentryALTinterwordspacing
L.~Simon, R.~Webster, and J.~Rabin, ``Revisiting precision and recall
  definition for generative model evaluation,'' \emph{arXiv preprint
  arXiv:1905.05441.}, 5 2019. [Online]. Available:
  \url{http://arxiv.org/abs/1905.05441}
\BIBentrySTDinterwordspacing

\bibitem{Roy2007}
O.~Roy and M.~Vetterli, ``The effective rank: A measure of effective
  dimensionality,'' in \emph{2007 15th European signal processing
  conference}.\hskip 1em plus 0.5em minus 0.4em\relax IEEE, 2007, pp. 606--610.

\bibitem{Zucker2010}
T.~A. Zucker, L.~M. Justice, S.~B. Piasta, and J.~N. Kaderavek, ``Preschool
  teachers' literal and inferential questions and children's responses during
  whole-class shared reading,'' \emph{Early Childhood Research Quarterly},
  vol.~25, pp. 65--83, 3 2010.

\bibitem{Habib2016}
\BIBentryALTinterwordspacing
M.~Habib, ``Assessment of reading comprehension,'' \emph{Revista Romaneasca
  pentru Educatie Multidimensionala}, vol.~8, pp. 125--147, 2016. [Online].
  Available: \url{http://dx.doi.org/10.18662/rrem/2016.0801.08}
\BIBentrySTDinterwordspacing

\end{thebibliography}
\end{document}